\documentclass[lettersize,journal]{IEEEtran}
\usepackage[T1]{fontenc}
\usepackage{amsmath,amsfonts}
\usepackage{algorithmic}
\usepackage{algorithm}
\usepackage{array}
\usepackage[caption=false,font=normalsize,labelfont=sf,textfont=sf]{subfig}
\usepackage{textcomp}
\usepackage{stfloats}
\usepackage{url}
\usepackage{verbatim}
\usepackage{graphicx}
\usepackage{cite}

\usepackage[dvipsnames]{xcolor}
\usepackage{booktabs}
\usepackage{multirow}
\usepackage{wrapfig}

\usepackage{makecell}
\usepackage[most]{tcolorbox}
\usepackage{caption}
\usepackage{balance}
\usepackage{colortbl} 

\tcbset{
    enhanced,
    sharp corners, %
    boxrule=0.4pt, %
    colback=white, %
    colframe=black, %
    fonttitle=\bfseries,
    fontupper=\normalsize, %
    left=6pt, right=6pt, top=4pt, bottom=4pt, %
}

\begin{document}

\title{3D-SceneQ: Empowering 3D LLM With Query-Guided Adaptive Pruning and Multi-Modal Feature Enhancement}

\author{Yongqi Shan, Lu Zhang,~\IEEEmembership{Member,~IEEE,} Jiazuo Yu, Yunzhi Zhuge, and Huchuan Lu,~\IEEEmembership{Fellow,~IEEE}
\thanks{\copyright~2026 IEEE. Personal use of this material is permitted. Permission from IEEE must be obtained for all other uses, in any current or future media, including reprinting/republishing this material for advertising or promotional purposes, creating new collective works, for resale or redistribution to servers or lists, or reuse of any copyrighted component of this work in other works. This is the author manuscript of an article published in \textit{IEEE Transactions on Multimedia}, vol.~28, pp.~5972--5983, 2026. DOI: 10.1109/TMM.2026.3668502.}
\thanks{This work was supported in part by the National Natural Science Foundation of China under Grants 62441231, 62406053 and 62206039, and in part by the Fundamental Research Funds for the Central Universities under Grants
DUT24YG119 and DUT24RC(3)025. \textit{(Corresponding author: Yunzhi Zhuge.)}}
\thanks{Yongqi Shan is with the School of Artificial Intelligence, Dalian University of Technology, Dalian 116081, China. (email: 11312327@mail.dlut.edu.cn)}
\thanks{Lu Zhang, Jiazuo Yu, Yunzhi Zhuge and Huchuan Lu are with the School of Information and Communication Engineering, Dalian University of Technology, Dalian 116081, China. (e-mail: zhangluu@dlut.edu.cn; yujiazuo@mail.dlut.edu.cn; zgyz@dlut.edu.cn; lhchuan@dlut.edu.cn)}
}

\maketitle

\begin{abstract}

Current 3D scene understanding pipelines typically concatenate vast numbers of 3D object tokens with text tokens and feed the resulting sequence to a Large Language Model (LLM). 
However, existing 3D-LLMs for scene understanding encounter two major limitations: (i) excessive inclusion of task-irrelevant object data, introducing noise that reduces reasoning accuracy and increases hallucinations; and (ii) reliance solely on point cloud data, which inherently lacks the rich semantic information available in complementary 2D modalities, such as color, material properties, texture, and high-level contextual relationships.
To address these challenges, we introduce 3D-SceneQ, a 3D LLM that unifies adaptive token pruning with multimodal semantic enrichment, markedly advancing scene understanding, reasoning, and grounding.
Specifically, we propose a Query-Guided Adaptive Pruning (QGAP) module that adaptively selects task-relevant objects based on the intended meaning of user instructions, leveraging a learnable latent query to retain critical information while effectively suppressing irrelevant noise.
In addition, we introduce a Multi-modal Object-level Feature Enhancement (MOFE) module, which integrates 2D feature embeddings from pre-trained models into 3D representation to enrich object-level semantic information. Evaluations show that 3D-SceneQ outperforms current state-of-the-art methods across multiple benchmarks (e.g., +5.7 CIDEr on Scan2Cap and +17.0 EM@1 on SQA3D), demonstrating its remarkable capabilities in various 3D language reasoning tasks.

\end{abstract}


\section{Introduction}
\label{sec:intro}

\IEEEPARstart{U}{nderstanding} 3D visual information is fundamental to numerous emerging technological domains, including autonomous navigation~\cite{arnold2019survey}, augmented and virtual reality~\cite{xiong2021augmented}, robotics~\cite{huang2023embodied}, and intelligent manufacturing systems~\cite{saimon2025advancing}. 
Effective interpretation of spatial structures and object relationships within 3D environments significantly enhances the capability of artificial intelligence (AI) systems to perceive, understand, and interact with the physical world. 
In recent years, considerable progress has been made in Large Language Models (LLMs), extending their applications from traditional natural language processing to advanced multimodal tasks, including complex 3D scene understanding~\cite{hong20223d,xiong20253ur}. 
Building upon these advancements, recent LLMs have demonstrated strong potential in 3D-related tasks. By integrating and analyzing rich spatial point cloud data, LLMs have demonstrated strong potential in a range of 3D-related tasks such as precise scene interpretation~\cite{he2024segpoint}, robust object perception~\cite{li20243dmit}, and advanced spatial reasoning~\cite{wang2023chat}. 
Such capabilities substantially enhance the contextual awareness and environmental comprehension of AI systems, ensuring more reliable and efficient performance in real-world applications. Therefore, continued exploration and development in this interdisciplinary field promise to foster deeper and more meaningful interactions among humans, machines, and complex physical environments.

\begin{figure}[t]
    \centering
    \includegraphics[width=\linewidth]{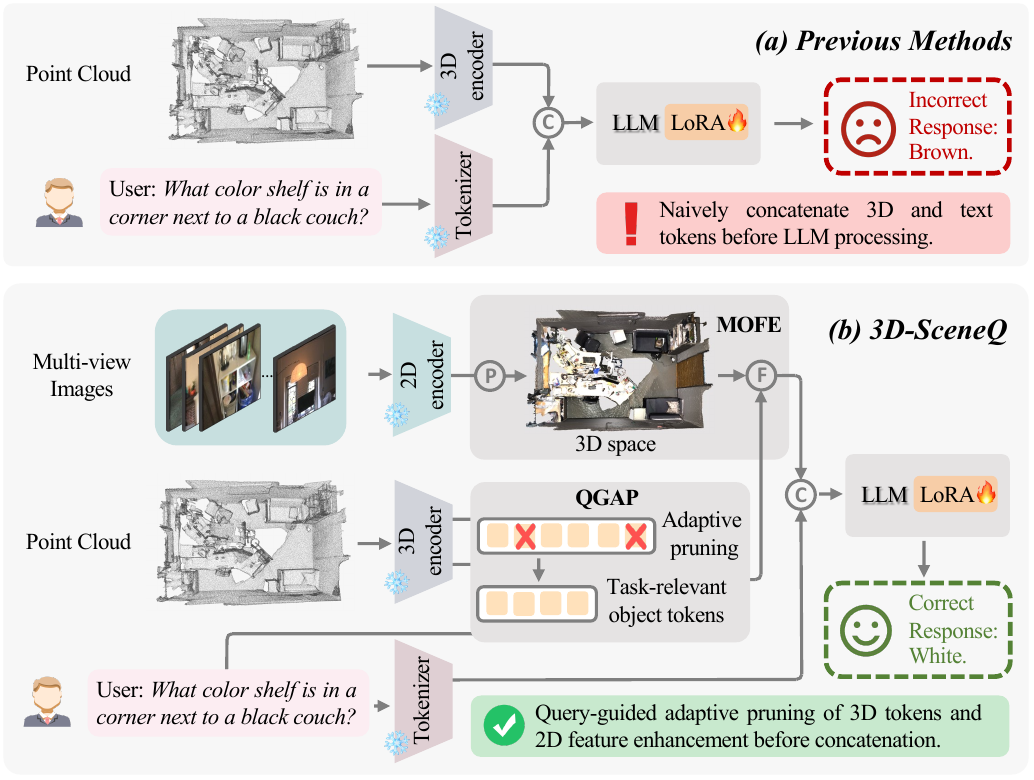} %
    \caption{
    (a) Conventional methods concatenate 3D and text tokens. (b) 3D-SceneQ uses a latent query to select task-relevant objects, then fuses 2D semantics for a compact, informative scene representation.
    \raisebox{-0.25ex}{\includegraphics[width=0.02\textwidth]{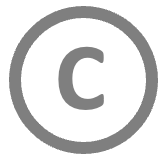}} represents the concatenation operation, \raisebox{-0.25ex} {\includegraphics[width=0.02\textwidth]{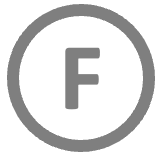}} denotes multimodal feature fusion, and \raisebox{-0.25ex} {\includegraphics[width=0.02\textwidth]{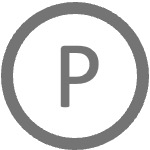}} denotes the projection of 2D features into 3D space.}
    \vspace{-3mm}
    \label{fig:compare}
\end{figure}
Although these methods have achieved significant advancements, comprehensive 3D scene understanding remains hindered by two major challenges:
(i) \textbf{Numerous Object Token Noise}: current 3D-LLMs~\cite{hong20233d,wang2023chat} typically employ 3D encoders (e.g. PointNet~\cite{qi2017pointnet} and PointNet++~\cite{qi2017pointnet++}) to convert 3D scenes into a large number of object tokens, among which only a small subset is task-relevant, introducing noise and increasing the likelihood of hallucinations~\cite{bang2023multitask}.
(ii) \textbf{Missing 2D Semantic Cues}: existing models~\cite{qi2017pointnet++,li20243dmit} for 3D scene understanding primarily rely on point cloud data, which inherently lacks rich semantic cues such as color, material, and high-level contextual relationships, all of which are typically provided by 2D visual features. 
However, most contemporary SOTA 3D LLMs, including Leo~\cite{huang2023embodied} and Point-LLM~\cite{xu2024pointllm} (see Fig.~\ref{fig:compare}(a)), ignore these two critical challenges and simply feed the LLM a concatenation of text and 3D object tokens, which limits the model's ability to reason about spatial semantics and to attain a nuanced understanding of complex 3D scenes.

To address these challenges, we introduce 3D-SceneQ, a novel 3D LLM designed to enhance scene perception, grounding, reasoning, and planning. As illustrated in Fig.~\ref{fig:compare}(b), 3D-SceneQ comprises two lightweight components: Query-Guided Adaptive Pruning (QGAP) and Multi-modal Object-level Feature Enhancement (MOFE).
\textbf{First}, QGAP tackles the challenge of task-relevant object selection by dynamically adjusting the number of objects retained based on the specific requirements of different questions. To achieve this adaptability, QGAP introduces a learnable latent query that adjusts the filtering ratio according to the question's informational needs. By tailoring object selection to the task, QGAP reduces noise, mitigates hallucinations, and enhances the model's effectiveness. \textbf{Second}, MOFE addresses the semantic limitations of point cloud data by incorporating 2D semantic features derived from multi-view scene images processed by pre-trained models such as CLIP~\cite{radford2021learning}. Exploiting the camera intrinsics and extrinsics, MOFE projects multi-view 2D features onto the 3D space, pools the resulting point-wise embeddings into object-level embeddings, and integrates them with 3D tokens through cross-attention to produce a unified multi-modal representation. Through object-level feature fusion, MOFE effectively integrates multi-modal information to enhance the semantic representation of task-relevant objects in 3D LLMs, enabling a richer and more comprehensive understanding of the scene. Unlike traditional multi-view stereo (MVS) methods~\cite{wang2021patchmatchnet,cabon2025must3r} that aim to reconstruct dense 3D geometry, or point-level contrastive approaches such as OpenScene~\cite{peng2023openscene}, MOFE focuses on injecting language-aligned semantic features into object-level 3D representations, enabling deeper multi-modal fusion without relying on geometric reconstruction.

Beyond their individual strengths, the synergistic integration of QGAP and MOFE gives rise to a virtuous cycle: QGAP filters out redundant clutter, sharpening the model's focus on task-relevant objects, while MOFE enriches these objects with multi-modal semantics, which in turn reinforces the relevance signals that guide QGAP's pruning. This synergy enables 3D-SceneQ to achieve a compact yet semantically rich understanding of complex 3D scenes, leading to consistently strong performance across diverse tasks. 
We evaluate 3D-SceneQ on a diverse set of 3D benchmarks, including 3D captioning~\cite{chen2021scan2cap}, 3D question answering~\cite{azuma2022scanqa}, and situated question answering~\cite{ma2022sqa3d}. 
Comprehensive experimental results show that 3D-SceneQ achieves state-of-the-art (SOTA) performance across all these tasks, achieving 78.1 CIDEr on Scan2Cap, 104.7 CIDEr on ScanQA, and 67.0 EM@1 on SQA3D, underscoring its strong generalization and reasoning capacity in complex 3D scenes. 
The main contributions of this work can be summarized as follows:

\begin{itemize} 
\item We introduce an innovative 3D-SceneQ to enhance 3D scene understanding and reasoning by reducing task-irrelevant information from 3D encoder and integrating 2D semantic features. Comprehensive experiments demonstrate that 3D-SceneQ achieves SOTA performance across diverse 3D understanding tasks.

\item We propose a Query-Guided Adaptive Pruning (QGAP) module, which dynamically selects task-relevant objects based on a specific user's question, mitigating hallucinations and improving reasoning accuracy in complex 3D environments by eliminating task-irrelevant tokens.

\item We develop a Multi-modal Object-level Feature
Enhancement (MOFE) module, which integrates semantically rich 2D features with spatially detailed 3D data to enrich 3D object-level semantic information, enabling deeper scene understanding and robust reasoning.

\end{itemize}

\section{Related works}
\subsection{Point Cloud Understanding}
Understanding 3D scenes is essential for intelligent systems to interact with the physical world. Unlike 2D images, 3D point clouds capture rich spatial and geometric information, making them crucial for tasks such as object recognition, manipulation, and reasoning in applications like robotics and autonomous driving. Early advances in 3D deep learning~\cite{yu2022point} were led by architectures such as PointGT~\cite{zhang2024pointgt} and DTNet~\cite{han2022dual}, which laid the foundation for 3D representation learning by effectively processing raw point clouds for classification and segmentation tasks. Meanwhile, inspired by vision-language models like CLIP~\cite{radford2021learning}, there has been growing interest in aligning 3D point clouds with language. Methods such as PointCLIP~\cite{zhang2022pointclip} and CLIP2Point~\cite{huang2023clip2point} project point clouds into 2D depth images for zero-shot classification using 2D CLIP models. Later works like ULIP~\cite{xue2023ulip} and OpenShape~\cite{liu2023openshape} employ contrastive learning to align 3D encoders with image and text through point-image-text triplets, improving the transferability of 3D representations. Unfortunately, these methods are inherently constrained to object-level tasks and struggle to generalize to scene-level understanding.

Recently, there has been growing interest in combining 3D point clouds with natural language to enable tasks~\cite{ye20223d, hong20233d, chen2020scanrefer, chen2021scan2cap, achlioptas2020referit3d}. 3D Question Answering~\cite{ye20223d} which focuses on answering questions grounded in 3D scenes, 3D Visual Grounding~\cite{chen2020scanrefer, chen2024grounded} which evaluates object localization from text and 3D captioning~\cite{chen2021scan2cap} further tests a model's ability to describe 3D scenes in natural language. However, most existing methods are designed for specific tasks and struggle to generalize beyond the distribution of their training data. In contrast, recent 3D Multi-modal Large Language Models (3D-LLMs) unify diverse 3D-language tasks within a single architecture, leveraging LLMs' reasoning abilities to handle QA, grounding, captioning, and more complex tasks like 3D dialogue and task decomposition. 

\subsection{Multi-Modal Large Language Models}
Multi-modal Large Language Models (MLLMs) are designed to understand and reason across diverse modalities beyond plain text, including images~\cite{li2022blip, li2023blip, radford2021learning, alayrac2022flamingo}, video~\cite{gong2025devil, xiong2025streaming}, and audio~\cite{rubenstein2023audiopalm}. These models aim to bridge the gap between language and perception by learning aligned representations across different input modalities. A landmark in this direction is CLIP~\cite{radford2021learning}, which jointly trains image and text encoders using contrastive learning on large-scale web data, enabling strong zero-shot transfer to downstream tasks. Building upon this, models such as BLIP~\cite{li2022blip} and BLIP-2~\cite{li2023blip} adopt a two-stage framework that connects frozen vision encoders with large language models via a learnable query transformer, followed by instruction tuning. LLaVA~\cite{liu2023visual} further enhances alignment by integrating CLIP visual features with open-source LLMs like Vicuna~\cite{chiang2023vicuna}, enabling multi-turn visual conversation and reasoning. More recent works, including Flamingo~\cite{alayrac2022flamingo}, Shikra~\cite{chen2023shikra}, and MiniGPT-4~\cite{zhu2023minigpt}, introduce gated cross-attention mechanisms and lightweight adapters to achieve more efficient and flexible vision-language integration. These models have shown strong performance across a variety of tasks such as visual question answering, image captioning, and multimodal instruction following, demonstrating the growing potential of MLLMs as general-purpose reasoning engines across modalities.

\begin{figure*}[t]
    \centering
    \includegraphics[width=\linewidth]{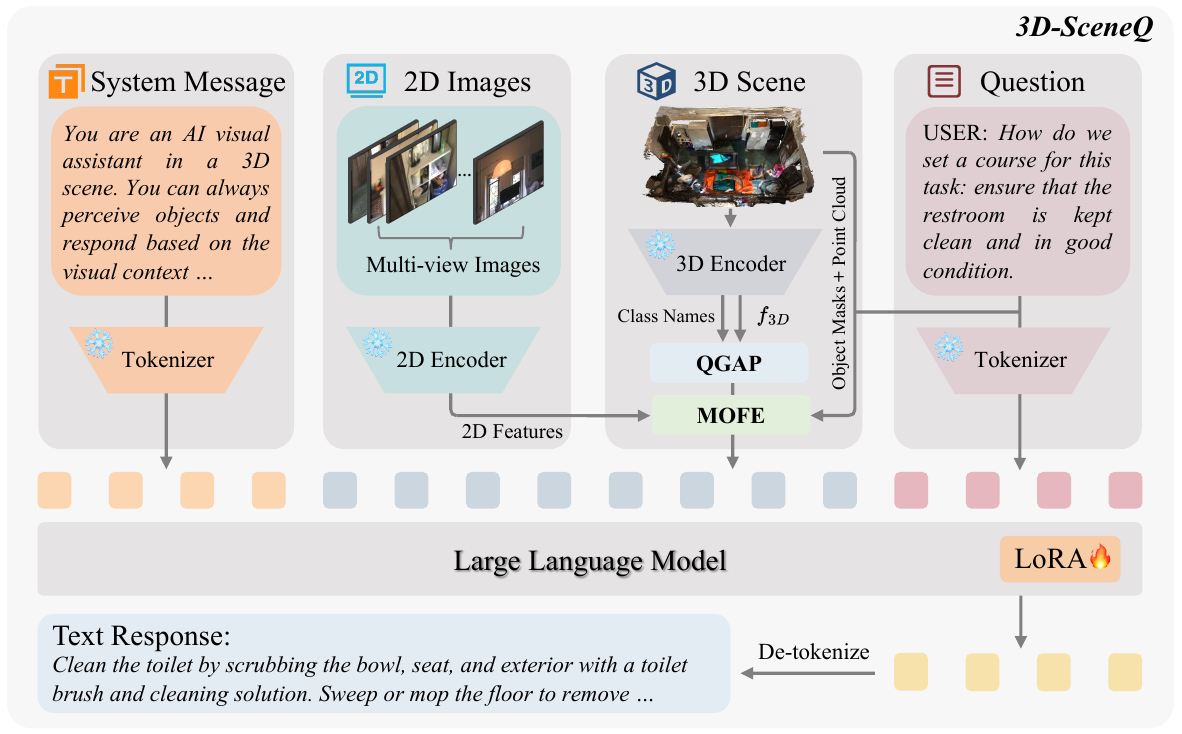} %
    \caption{Overview of the 3D-SceneQ framework, which integrates multi-view 2D images, 3D point clouds, and textual instructions to generate informed responses. QGAP refers to the Query-Guided Adaptive Pruning module, and MOFE denotes the Multi-modal Object-level Feature Enhancement module.}
    \vspace{-2mm}
    \label{fig:1}
\end{figure*}

\subsection{3D Multi-Modal Large Language Models} 
The recent success of multi-modal large language models (LLMs) has spurred growing interest in extending their capabilities to the 3D domain~\cite{hong20233d, shi2025multi, he2024segpoint, xu2024pointllm, li20243dmit, han2024onellm, guo2023point, qi2024gpt4point, huang2023embodied, yin2025shapegpt, wu2025language}. Among these efforts, 3D-LLM~\cite{hong20233d} introduces 3D point cloud features into pre-trained 2D vision-language models (VLMs), enabling spatially grounded understanding across diverse 3D-language tasks. SegPoint~\cite{he2024segpoint} adopts a unified architecture that directly feeds point cloud data into LLMs for a wide range of segmentation tasks. Point-LLM~\cite{xu2024pointllm} employs a two-stage training approach---first aligning 3D features with the language space, followed by instruction tuning---which facilitates stronger integration between geometric understanding and linguistic reasoning. In contrast, 3D-MiT~\cite{li20243dmit} proposes a more streamlined framework by eliminating the explicit alignment stage, directly fusing visual features and textual prompts to construct multimodal 3D inputs with reduced training complexity. OneLLM~\cite{han2024onellm} and Point-Bind LLM~\cite{guo2023point} further explore the alignment of 3D point clouds with other modalities---including images, audio, and text---within a shared embedding space, thereby enhancing cross-modal reasoning and enabling seamless interaction between geometric and semantic cues. GPT4Point~\cite{qi2024gpt4point} extends this direction with a generative multimodal model that supports 3D understanding and controllable output generation. LEO~\cite{huang2023embodied} represents a different line of work, focusing on embodied agents that integrate 3D point clouds and egocentric RGB images for manipulation tasks. However, in LEO, visual inputs are primarily used as low-level control signals, without integrating the high-level semantics of images into a unified 3D representation.

Despite these advancements, a fundamental challenge remains: point clouds provide rich geometric information but are inherently sparse in semantic content, and often contain significant structural redundancy. Most existing architectures do not explicitly model these characteristics, limiting their ability to fully leverage the expressive potential of 3D representations for downstream reasoning and decision-making tasks.

\section{Methodology}
\label{sec:method}
\subsection{3D-SceneQ}
3D-SceneQ adopts a two-stage learning scheme to integrate multi-modal information for downstream tasks: (i) 3D Vision-Language Alignment, aligning 3D object features with linguistic representations to form a shared feature space; and (ii) Instruction Tuning, fine-tuning the model with task-specific instructions to improve 3D understanding and reasoning. In the first stage, we employ the object-centric 3D point cloud encoder, PointNet++~\cite{qi2017pointnet++}, to interpret the scene. This encoder extracts geometric features for each 3D object and represents them as point cloud embeddings. Spatial Transformer~\cite{chen2022language} is then applied after feature extraction to process the point cloud embeddings of all objects, transforming them into object-centric 3D object tokens. Subsequently, these 3D object tokens are projected through a linear layer to align their dimensionality with the requirements of the LLM, producing the final 3D object representation $f_{3D} \in \mathbb{R}^{N \times d}$, where $N$ is the number of objects in the scene, and $d$ is the feature dimensionality compatible with the LLM. The input in this stage includes system messages, user's questions, and 3D object tokens $f_{3D}$, forming a multi-modal sequence. This sequence is then processed by the pre-trained LLM to generate the final response. Text-based inputs, such as system messages and user's questions, are tokenized using SentencePiece~\cite{kudo2018sentencepiece} with a 32k vocabulary.

The second stage applies instruction tuning for 3D task, it builds upon the first stage but introduces additional mechanisms to enhance task-specific performance. As shown in Fig.~\ref{fig:1}, the 3D features are refined through Query-Guided Adaptive Pruning (QGAP) and enriched using Multi-modal Object-level Feature Enhancement (MOFE). Details of QGAP and MOFE are provided in Section~\ref{sec:QGAP} and Section~\ref{sec:MOFE}, respectively. These two processes result in enhanced 3D object tokens \( f_{E_{3D}} \in \mathbb{R}^{K \times d} \), where \( K \) represents the number of objects remaining in the scene after pruning by QGAP. The input multi-modal sequence in this stage includes system messages, questions and enhanced 3D object tokens \( f_{E_{3D}}\). For a batch \( \mathcal{B} \) of token sequences, 3D-SceneQ is optimized using the following objective function $\mathcal{L}$:
\begin{equation}
\mathcal{L}(\theta, \mathcal{B}) = - \sum_{b=1}^{|\mathcal{B}|} \sum_{t=1}^{T} \log p_\theta \left( s_{\text{res}}^{(b,t)} \mid s_{\text{res}}^{(b,<t)}, s_{\text{prefix}}^{(b)} \right),
\end{equation}
where \( s_{\text{res}}^{(b,t)} \) denotes the target token at time step \( t \), \( s_{\text{res}}^{(b,<t)} \) represents all preceding tokens in the response, and \( s_{\text{prefix}}^{(b)} \) comprises the system message, 2D and 3D features, and the input question. \( T \) denotes the length of the target response sequence. The probability distribution over the next token is parameterized by \( p_\theta \), where \( \theta \) denotes the model parameters.

\subsection{Query-Guided Adaptive Pruning}\label{sec:QGAP}
\begin{figure}[t]
    \centering
    \includegraphics[width=\linewidth]{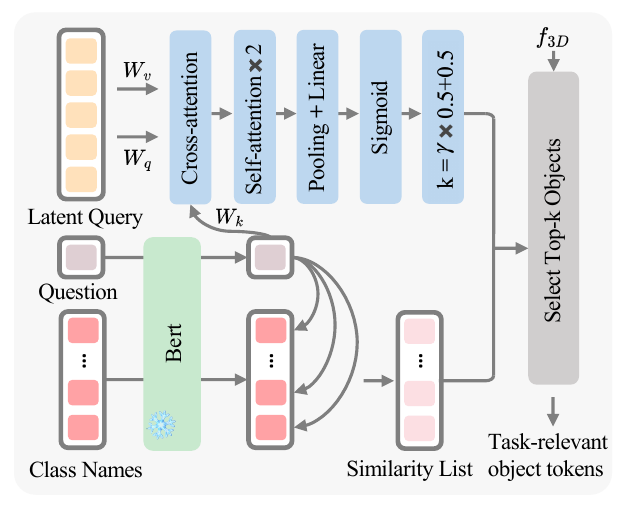} %
    \caption{Overview of the Query-Guided Adaptive Pruning (QGAP) module, which retains task-relevant object tokens based on a global latent query.}
    \label{fig:qgap}
\end{figure}

To adaptively filter object tokens based on relevance to a specific task, we propose Query-Guided Adaptive Pruning (QGAP). 
It dynamically retains the most relevant object tokens by comparing the user's instruction with predicted 3D object category labels.
As presented in Fig.~\ref{fig:qgap}, given a scene's point cloud \( p \) and a question \( q \), we encode \( q \) using a pre-trained, frozen BERT~\cite{devlin2018bert} to obtain the question embedding \( q_{\text{emb}} \):
\begin{equation}
q_{\mathrm{emb}}=\operatorname{BERT}(q).
\end{equation}

Next, the scene's point cloud \( p \) is processed by a 3D encoder PointNet++ to extract object tokens \( o_i \) and their corresponding predicted categories $\text{label}_i$. Each predicted category is then transformed into an embedding \( o_{\text{emb}}^{(i)} \) using BERT~\cite{devlin2018bert}:
\begin{equation}
o_{\text{emb}}^{(i)} = \text{BERT}(\text{label}_i), \quad i = 1, \ldots, n,
\end{equation}
where \( n \) denotes the total number of object tokens in the scene. We further compute the cosine similarity between \( q_{\text{emb}} \) and each \( o_{\text{emb}}^{(i)} \) to assess the relevance of each object token:
\begin{equation}
\text{sim}(q_{\text{emb}}, o_{\text{emb}}^{(i)}) = \frac{q_{\text{emb}} \cdot o_{\text{emb}}^{(i)}}{\|q_{\text{emb}}\| \|o_{\text{emb}}^{(i)}\|},
\end{equation}
these similarity scores are used to rank the objects by their relevance to \( q_{\text{emb}} \) and are organized into a sorted list $\boldsymbol{S}$: 
\begin{equation}
\boldsymbol{S} = \{\text{sim}(q_{\text{emb}}, o_{\text{emb}}^{(i)})\}.
\end{equation}

To further adaptively prune the scene's object tokens based on the similarity scores stored in $\boldsymbol{S}$, we introduce a globally learnable latent query $LQ$. The latent query $LQ$ captures task-relevant information by interacting with the question embedding $q_{\text{emb}}$ through a cross-attention mechanism. To further refine the extracted features, the cross-attention output is processed through two layers of self-attention. Subsequently, the refined features are aggregated along the sequence dimension via a pooling operation. The aggregated representation is passed through a linear transformation followed by a sigmoid activation function to compute the initial filtering ratio $\gamma$:
\begin{equation}
\gamma = \sigma(\text{Linear}(\text{Pool}(\text{SelfAttn}^2(\text{CrossAttn}(LQ, q_{\text{emb}}))))),
\end{equation}
where \(\sigma\) denotes the sigmoid activation function. To ensure a reasonable proportion of objects is retained, the final number of object tokens $k$ is determined as follows:
\begin{equation}
k = \text{round} \left( \left( \frac{\gamma}{2} + 0.5 \right) \times n \right).
\end{equation}

Finally, we select the top-$k$ object tokens with the highest relevance to the question based on the similarity list $\boldsymbol{S}$. These tokens are then retained for downstream processing.

\subsection{Multi-modal Object-level Feature Enhancement}\label{sec:MOFE}
\begin{figure*}[t]
    \vspace{-1mm}
    \centering
    \includegraphics[width=1\linewidth]{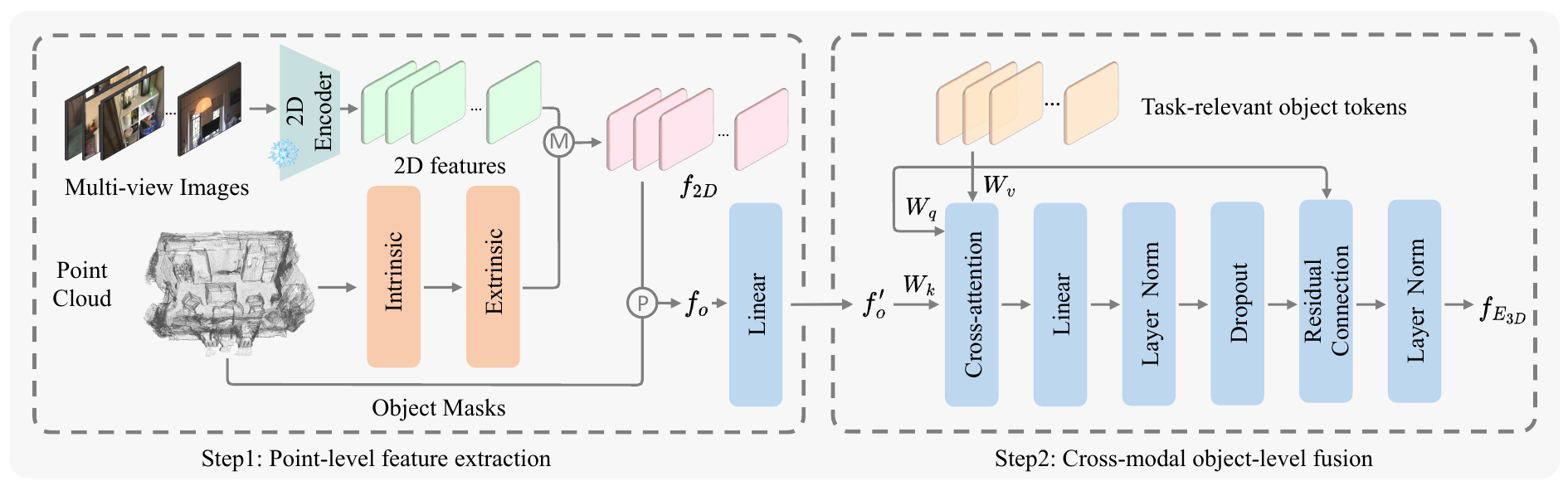} %
    \vspace{-1mm}
    \caption{Overview of the Multi-modal Object-level Feature Enhancement (MOFE) module, which combines features across modalities. \raisebox{-0.375ex}{\includegraphics[width=0.02\textwidth]{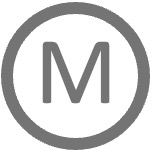}} represents mapping multi-view 2D features to each point in 3D space, with aggregation and averaging across views. \raisebox{-0.375ex}{\includegraphics[width=0.02\textwidth]{pictures/p.jpg}} denotes pooling point-level features into object-level features using object masks.}
    \vspace{-3mm}
    \label{fig:3}
\end{figure*}
Multi-modal Object-level Feature Enhancement (MOFE) consists of two main steps: Point-level Feature Extraction and Cross-modal Object-level Fusion. As shown in Fig.~\ref{fig:3}, we extract dense, per-pixel embeddings from multi-view images and project them onto the 3D surface to integrate 2D image features with 3D point cloud data. These projected features are then fused into a unified multi-modal representation. 

\textbf{Point-level feature extraction.} For an RGB image with resolution \( H \times W \), we use the pre-trained segmentation model OpenSeg~\cite{ghiasi2022scaling} to extract per-pixel embeddings \( f_i \in \mathbb{R}^{H \times W \times C} \), where \( C \) is the embedding dimension and \( i \) denotes the image index in a sequence of views. Each 3D point \( p \in \mathbb{R}^3 \) in the point cloud \( P \in \mathbb{R}^{M \times 3} \), with \( M \) points, is projected onto its corresponding pixel \( u = (x, y) \) in the image using the camera's intrinsic \( I_i \) and extrinsic \( E_i \) matrices in the image for each view. This process can be expressed as:
\begin{equation}
u = I_i \cdot E_i \cdot p.
\end{equation}

To aggregate features from multiple views for each projected point \( p \), we average the per-pixel embedding from each view at the corresponding pixel location \( u \), as expressed by:
\begin{equation}
f_{2D}(u) = \frac{1}{V} \sum_{i=1}^{V} f_i(u),
\end{equation}
where \( V \) is the number of views. This results in a set of point-level 2D features \( f_{2D} \in \mathbb{R}^{M \times C} \), where each row corresponds to the aggregated feature of a point in the point cloud.

\textbf{Cross-modal object-level fusion.} 
We first employ object masks \( M_o \) to identify the sets of points belonging to each object in the point cloud. By pooling the 2D features \( f_{2D} \) associated with all points within each object region, we derive an object-level feature vector \( f_o \). To align this object-level representation with the input requirements of the LLM, we apply a linear projection parameterized by \( W_{\text{proj}} \in \mathbb{R}^{d \times C} \), which maps \( f_o \) into a \( d \)-dimensional space. Formally, this transformation can be expressed as:
\begin{equation}
f_{o}^{\prime} = W_{\text{proj}} \cdot f_{o},
\end{equation}
where \( f_{o}^{\prime} \) is the final object-level feature vector compatible with the LLM input dimension. We perform cross-modal fusion using cross-attention, where 3D object-level features, extracted by a pre-trained 3D encoder, serve as queries \( Q \) and values \( V \), transformed 2D object-level features provide keys \( K \). This process is defined as: 
\begin{equation}
f_{fuse} = \text{softmax}\left(\frac{Q K^T}{\sqrt{d_k}}\right) V,
\end{equation}
where $Q = W_q \cdot f_{3D}$, $K = W_k \cdot f_{o}^{\prime}$, $V = W_v \cdot f_{3D}$. $W_q$, $W_k$ and $W_v \in \mathbb{R}^{d_k \times d}$ are learnable projection matrices. $f_{3D} \in \mathbb{R}^{N \times d}$ represents 3D object-level features extracted by the pre-trained 3D encoder. The parameter $d_k$ scales the attention scores. To enhance robustness, dropout and a residual connection are applied, resulting in the fused feature $f_{E_{3D}}$, which is then fed into the LLM for multi-modal 3D scene understanding:
\begin{equation}
f_{E_{3D}} = \text{Dropout}(f_{fuse}) + f_{3D}.
\end{equation}

\section{Expirements}
\label{sec:expirements}

\subsection{Experimental Setups}
\textbf{Benchmarks.} Following~\cite{huang2023embodied}, we utilise a diverse set of 3D vision-language datasets tailored for alignment and instruction tuning. In the 3D vision-language alignment phase, we train the model to bridge the gap between 3D scene representations and natural language using three types of captioning data. These include (i) object-level captions from \texttt{Cap3D}\cite{luo2024scalable}, which provide fine-grained textual descriptions of objects from \texttt{Objaverse}\cite{deitke2023objaverse}; (ii) object-in-context referring expressions from \texttt{ScanScribe}\cite{zhu20233d} and \texttt{ReferIt3D}\cite{achlioptas2020referit3d}, further augmented with LLM-generated annotations to enhance contextual diversity; and (iii) scene-level summaries from \texttt{3RScan}\cite{wald2019rio}, capturing global information about object layouts, spatial relations, and scene semantics. 
In the instruction tuning phase, we refine the model's ability to understand, reason, and respond to natural language instructions in 3D environments. We utilize tasks based on \texttt{ScanNet}\cite{dai2017scannet} and \texttt{3RScan}, covering open-ended 3D captioning and question answering, where the model is expected to describe scenes or answer queries about object properties, locations, and relationships. 

\textbf{Evaluation Metrics.}  We use a set of standard evaluation metrics to assess model performance across various tasks. For language generation tasks, we report CIDEr, BLEU, METEOR, and ROUGE, which measure relevance, fluency, and overlap with reference texts from different perspectives. 
In open-ended generation tasks, where multiple correct answers may exist for a given input, we additionally use sentence similarity~\cite{reimers2019sentence} to better capture semantic alignment. 
For question-answering tasks that require factual precision, we report top-1 exact match accuracy (EM@1), which reflects the proportion of predictions that exactly match the ground truth. 

\begin{table}[t]
\vspace{2mm}
\definecolor{explainclr}{rgb}{0.6,0.725,0.650} 
\newcommand{\prompttxt}[1]{#1}  %
\newcommand{\explaintext}[1]{\textcolor{explainclr}{\texttt{#1}}}

\begin{tcolorbox}[
    colback=gray!10,
    colframe=black,
    boxrule=0.5pt,
    arc=3pt,
    left=6pt, right=3pt, top=6pt, bottom=6pt,
    width=\linewidth,
    fontupper=\small
]

\textbf{Role Prompt:}

\prompttxt{You are an AI assistant situated in a 3D environment. You perceive all scene objects, including yourself, and respond to user instructions based on visual input.}

\vspace{0.05em}
\rule{\linewidth}{0.4pt}

\textbf{3D Objects Prompt:}

\prompttxt{Objects (including you) in the scene:}

\explaintext{This section is followed by 3D scene features extracted by a pre-trained encoder, capturing object properties and spatial relations.}

\vspace{0.05em}
\rule{\linewidth}{0.4pt}

\textbf{Task Prompt:}

\prompttxt{User: \texttt{\textit{\{instruction\}}} ASSISTANT: }

\explaintext{The placeholder \textit{\{instruction\}} is replaced with the user's natural language query during inference.}

\end{tcolorbox}
\vspace{-1mm}
\captionof{table}{Prompt structure used in 3D-SceneQ, combining role description, object context, and user instruction in a unified input format.}
\label{tab:3dsceneq-promptbox}

\setlength{\parindent}{1em}  %
\end{table}

\begin{table}[t]
    \centering
    \vspace{-1mm}
    \begin{tabular}{l c}
    \toprule
    \textbf{Component} & \textbf{Number of Parameters} \\
    \midrule
    QGAP & 7.1M \\
    MOFE & 70.3M \\
    Trainable Parameters (Total) & 122.6M \\
    3D-SceneQ (Overall) & 7.0B \\
    \bottomrule
    \end{tabular}
    \caption{Parameter overview for 3D-SceneQ and its key components.}
    \label{tab:3d-sceneq-params}
    \vspace{-4mm}
\end{table}

\textbf{Implementation Details.}  We adopt Vicuna-7B~\cite{chiang2023vicuna} as the language backbone, and we directly utilize the pre-trained model from the first stage of LEO~\cite{huang2023embodied} for initialization. 
The 3D encoder, the 2D semantic segmentation model used in MOFE, the language tokenizer and language backbone, as well as the BERT encoders used for query and object category representation in QGAP, are kept frozen during training. 
We adopt LoRA~\cite{hu2021lora} with a low-rank setting ($r=16$) to enable parameter-efficient fine-tuning of the Vicuna-7B language backbone.
The remaining trainable parameters, including the proposed QGAP, MOFE, the object-to-text feature mapping layer, and the spatial transformer built atop PointNet++, are trained from scratch. 
Training is conducted for 10 epochs using the AdamW optimizer with a learning rate of $3\times10^{-5}$.
All experiments are performed on two NVIDIA A800 GPUs.

We propose a prompt structure to facilitate 3D-SceneQ in comprehending tasks by integrating the model's role description, object context within the scene, and user instructions, as detailed in Table~\ref{tab:3dsceneq-promptbox}. In terms of model size, 3D-SceneQ comprises approximately 7 billion parameters, yet only 122.6 million parameters are tunable during training, as summarized in Table~\ref{tab:3d-sceneq-params}. Notably, the QGAP and MOFE modules are exceptionally lightweight, contributing merely 7.1 million and 70.3 million parameters, respectively. As a result, these modules have a negligible impact on the overall computational and memory footprint of the model.

\renewcommand{\arraystretch}{1.1}
\vspace{-1mm}
\begin{table*}[t]
\centering
\caption{Comparison of 3D-SceneQ with state-of-the-art models on Scan2Cap, ScanQA, and SQA3D benchmarks across various evaluation metrics. ``EM@1" denotes top-1 exact match accuracy; Scan2Cap n-gram metrics use IoU $\ge$ 0.5. \textsuperscript{\textdagger} is GPT-3 answers based on generated captions. ``Sim'' denotes sentence similarity.}
\label{tab:results}
\resizebox{0.95\linewidth}{!}{
\begin{tabular}{l>{\centering\arraybackslash}p{0.97cm}>{\centering\arraybackslash}p{0.97cm}>{\centering\arraybackslash}p{0.97cm}>{\centering\arraybackslash}p{0.97cm}>{\centering\arraybackslash}p{0.97cm}>{\centering\arraybackslash}p{0.97cm}>{\centering\arraybackslash}p{0.97cm}>{\centering\arraybackslash}p{0.97cm}>{\centering\arraybackslash}p{0.97cm}>{\centering\arraybackslash}p{1.5cm}}
\toprule
\multirow{3.5}{*}{Method} & \multicolumn{5}{c}{\raisebox{-1.5ex}{Scan2Cap (val)}} & \multicolumn{4}{c}{\raisebox{-1.5ex}{ScanQA (val)}} & SQA3D (test) \\
\cmidrule(lr){2-6} \cmidrule(lr){7-10} \cmidrule(lr){11-11}
& CIDEr & BLEU-4 & METEOR & ROUGE & Sim & CIDEr & BLEU-4 & METEOR & ROUGE & EM@1 \\
\midrule
\textit{Task-specific models} \\
Scan2Cap~\cite{chen2021scan2cap} & 35.2 & 22.4 & 21.4 & 43.5 & - & - & - & - & - & 41.0\textsuperscript{\textdagger} \\
Vote2Cap-DETR~\cite{chen2023end} & 61.8 & 34.5 & 26.2 & 54.4 & - & - & - & - & - & - \\
ScanRefer+MCAN~\cite{yu2019deep} & - & - & - & - & - & 55.4 & 7.9 & 11.5 & 30.0 & - \\
ScanQA~\cite{azuma2022scanqa} & - & - & - & - & - & 64.9 & 10.1 & 13.1 & 33.3 & 47.2 \\
\midrule
\textit{Task-specific fine-tuned} \\
3DMIT~\cite{li20243dmit}& - & - & - & - & - & 48.0 & 5.2 & 10.7 & 26.2 & - \\ 
Chat-3D~\cite{wang2023chat}& - & - & - & - & - & 53.2 & 6.4 & 11.9 & 28.5 & - \\
3D-VisTA~\cite{zhu20233d} & 66.9 & 34.0 & 27.1 & 54.3 & 53.8 & 69.6 & 10.4 & 13.9 & 35.7 & 48.5 \\
3D-LLM (FlanT5)~\cite{hong20233d} & - & - & - & - & - & 69.4 & 12.0 & 14.5 & 35.7 & - \\
NaviLLM~\cite{zheng2024towards}& - & - & - & - & - & 75.9 & 12.5 & 15.4 & 38.4 & - \\
LEO~\cite{huang2023embodied} & 72.4 & 38.2 & 27.9 & 58.1 & 55.3 & 101.4 & 13.2 & 20.0 & \textbf{49.2} & 50.0\\
\midrule
3D-SceneQ & \textbf{78.1} & \textbf{43.6} & \textbf{30.5} & \textbf{65.3} & \textbf{64.2} & \textbf{104.7} & \textbf{13.6} & \textbf{20.4} & 49.1 & \textbf{67.0} \\
\bottomrule
\end{tabular}}
\end{table*}

\renewcommand{\arraystretch}{1.1}
\begin{table*}[t]
\vspace{-1mm}
\centering
{\setlength{\tabcolsep}{6pt}
\caption{Performance of 3D-SceneQ with different pruning and enhancement strategies. RandFilter refers to random pruning without adaptive mechanisms.}
\label{tab:ablations}
\resizebox{0.95\linewidth}{!}{
\begin{tabular}{l>{\centering\arraybackslash}p{1cm}>{\centering\arraybackslash}p{1cm}>{\centering\arraybackslash}p{1cm}>{\centering\arraybackslash}p{1cm}>{\centering\arraybackslash}p{1cm}>{\centering\arraybackslash}p{1cm}>{\centering\arraybackslash}p{1cm}>{\centering\arraybackslash}p{1cm}>{\centering\arraybackslash}p{1cm}>{\centering\arraybackslash}p{1.5cm}}
\toprule
\multirow{3.5}{*}{Method}& \multicolumn{5}{c}{\raisebox{-1.5ex}{Scan2Cap (val)}} & \multicolumn{4}{c}{\raisebox{-1.5ex}{ScanQA (val)}} & SQA3D (test) \\
\cmidrule(lr){2-6} \cmidrule(lr){7-10} \cmidrule(lr){11-11}
& CIDEr & BLEU-4 & METEOR & ROUGE & Sim & CIDEr & BLEU-4 & METEOR & ROUGE & EM@1 \\
\midrule
Baseline & 72.4 & 38.2 & 27.9 & 58.1 & 55.3 & 101.4 & 13.2 & 20.0 & \textbf{49.2} & 50.0 \\
Baseline + RandFilter & 52.3 & 38.0 & 27.8 & 62.3 & 55.2 & 90.1 & 13.4 & 18.1 & 43.7 & 68.9 \\
Baseline + QGAP & \textbf{78.4}& 44.1& 30.4& 65.6 & 64.2& 101.6& 14.1& 19.9& 48.7& \textbf{70.1}\\
Baseline + MOFE & 77.8 & \textbf{44.4} & 30.5 & 65.1 & 64.1 & 102.0 & 14.6 & 20.0 & 48.8 & 66.7 \\
3D-SceneQ & 78.1 & 43.6 & \textbf{30.5} & \textbf{65.3} & \textbf{64.2} & \textbf{104.7} & \textbf{15.2} & \textbf{20.4} & 49.1 & 67.0 \\
\bottomrule
\end{tabular}
}}
\vspace{-3mm}
\end{table*}

\subsection{Comparison with State-of-the-Arts}

We evaluate our method across three prominent 3D scene understanding tasks: 3D scene captioning with Scan2Cap~\cite{chen2021scan2cap}, 3D question answering (QA) with ScanQA~\cite{azuma2022scanqa}, and 3D scene-based reasoning with SQA3D~\cite{ma2022sqa3d}. As summarized in Table~\ref{tab:results}, 3D-SceneQ consistently achieves state-of-the-art performance on all these benchmarks, surpassing strong task-specific baselines as well as fine-tuned specialized models across all evaluation metrics considered. 
Specifically, on Scan2Cap it lifts METEOR to 30.5 and ROUGE to 65.3, exceeding the previous best by 2.6 and 7.2 points, respectively; on ScanQA it pushes CIDEr to 104.7 and BLEU-4 to 13.6, reflecting more relevant and fluent answers; and on the SQA3D test set it attains an EM@1 of 67.0\%, a striking 17 point leap over the strongest prior method, demonstrating exceptional proficiency in generating precise and contextually informative outputs. 
In contrast to conventional methods that depend on carefully engineered task-specific architectures, 3D-SceneQ adopts a unified LLM-based framework augmented with lightweight components---QGAP and MOFE. This unified architecture allows the model to flexibly and effectively adapt to diverse 3D reasoning tasks, enabling robust open-ended response generation while maintaining highly competitive quantitative performance achieved by specialized methods.

\subsection{Ablation Studies}
\subsubsection{Effectiveness of Core Components}
We conducted ablation studies in Table~\ref{tab:ablations} to disentangle the individual and combined effects of 3D-SceneQ's two core modules: Query-Guided Adaptive Pruning (QGAP) and the Multi-modal Object-Feature Enhancer (MOFE). Using the LEO~\cite{huang2023embodied} architecture with embodied agent related components removed as the Baseline model, we draw four nuanced observations: 
\textbf{First, random pruning helps reduce redundancy but risks discarding crucial cues.}  
The RandFilter variant stochastically discards 40\% of object tokens whenever a scene contains more than 20 objects. This na{\"i}ve compression alone propels SQA3D EM@1 from 50.0 to 68.9 by reducing extraneous information evidence that minimizing redundant information helps the model reason more effectively. Yet the indiscriminate pruning also removes salient objects and attributes, collapsing Scan2Cap CIDEr from 72.4 to 52.3 and diminishing ScanQA performance as well. The result exposes a core tension: token economy is vital, but pruning must be relevance-aware.
\textbf{Second, QGAP enhances both descriptive quality and reasoning accuracy.}  
QGAP learns a filtering ratio over object tokens conditioned on the user's question, excising only those entities that contribute negligibly to the downstream objective. This targeted sparsification raises Scan2Cap CIDEr to 78.4 and SQA3D EM@1 to 70.1, outperforming both the Baseline and RandFilter. The gains affirm that relevance-aware pruning preserves key geometric and semantic cues while still reaping the efficiency benefits of a lighter context.
\textbf{Third, MOFE boosts holistic performance, especially on ScanQA.}  
By fusing geometric, visual, and linguistic signals into a refined object embedding, MOFE elevates Scan2Cap CIDEr to 77.8 and boosts SQA3D EM@1 to 66.7. The improvement on ScanQA is particularly pronounced, suggesting that MOFE's cross-modal representations equip the model to resolve subtle spatial references and attribute dependencies that commonly confound 3D question answering.
\textbf{Finally, synergistic coupling of QGAP and MOFE yields the most balanced, robust gains.}  
Our full model, 3D-SceneQ, combines QGAP's task-relevant pruning strategy with MOFE's semantic enrichment, producing consistent improvements across all benchmarks (e.g., +5.7 CIDEr over Baseline on Scan2Cap, +20.0 EM@1 on SQA3D). 
Guided by the user query, QGAP selects task-relevant object tokens. MOFE then enriches these tokens with 2D features and feeds back sharper relevance cues to QGAP. This virtuous loop enables 3D-SceneQ to form a compact yet information-dense representation of complex 3D scenes, yielding balanced gains across diverse tasks.

\renewcommand{\arraystretch}{1}
\begin{table*}[h]
\vspace{-1mm}
\centering
{\setlength{\tabcolsep}{5pt}
\caption{Performance of 3D-SceneQ with different token pruning ratios. ``LowSimFilt" refers to pruning a fixed proportion of object tokens with the lowest similarity to the query.}
\label{tab:filt1}
\resizebox{1\linewidth}{!}{
\begin{tabular}{l>{\centering\arraybackslash}p{1.4cm}>{\centering\arraybackslash}p{1.4cm}>{\centering\arraybackslash}p{1.4cm}>{\centering\arraybackslash}p{1.4cm}>{\centering\arraybackslash}p{1.4cm}>{\centering\arraybackslash}p{1.4cm}>{\centering\arraybackslash}p{1.4cm}>{\centering\arraybackslash}p{1.4cm}>{\centering\arraybackslash}p{1.4cm}>{\centering\arraybackslash}p{1.4cm}}
\toprule
\multirow{3.5}{*}{Method}& \multicolumn{5}{c}{\raisebox{-1.5ex}{Scan2Cap (val)}} & \multicolumn{4}{c}{\raisebox{-1.5ex}{ScanQA (val)}} & SQA3D (test) \\
\cmidrule(lr){2-6} \cmidrule(lr){7-10} \cmidrule(lr){11-11}
& CIDEr & BLEU-4 & METEOR & ROUGE & Sim & CIDEr & BLEU-4 & METEOR & ROUGE & EM@1 \\
\midrule
Baseline & 72.4 & 38.2 & 27.9 & 58.1 & 55.3 & 101.4 & 13.2 & 20.0 & \textbf{49.2} & 50.0 \\
Baseline +10\% LowSimFilt &  73.7&  43.7&  30.0&  65.2&  63.1&  93.3&  13.2&  18.3&  45.6&  67.2\\
Baseline +20\% LowSimFilt &  74.5&  43.5&  30.2&  65.3&  63.8&  97.6&  13.2&  18.8&  46.7&  69.0\\
Baseline +30\% LowSimFilt &  69.8&  42.7&  29.7&  64.8&  62.0&  90.5&  12.4&  18.0&  44.5&  69.0\\
Baseline +40\% LowSimFilt &  67.2&  42.4&  29.6&  64.3&  61.9&  89.4&  11.5&  16.5&  42.4&  61.3\\
Baseline +50\% LowSimFilt &  66.0&  41.8&  29.4&  64.0&  61.5&  81.5&  10.5&  16.4&  41.4&  59.5\\
Baseline + QGAP  &  \textbf{78.4}& \textbf{44.1}& \textbf{30.4}& \textbf{65.6}& \textbf{64.2}& \textbf{101.6}& \textbf{14.1}& \textbf{19.9}& 48.7& \textbf{70.1}\\
\bottomrule
\end{tabular}
}}
\end{table*}

\renewcommand{\arraystretch}{1}
\begin{table*}[h]
\vspace{-1mm}
\centering
{\setlength{\tabcolsep}{6pt}}
\caption{Performance of 3D-SceneQ with different token pruning ratios. ``RandFilt" denotes randomly removing a fixed percentage of object tokens without adaptive selection.}
\label{tab:filt2}
\resizebox{1\linewidth}{!}{
\begin{tabular}{l>{\centering\arraybackslash}p{1.4cm}>{\centering\arraybackslash}p{1.4cm}>{\centering\arraybackslash}p{1.4cm}>{\centering\arraybackslash}p{1.4cm}>{\centering\arraybackslash}p{1.4cm}>{\centering\arraybackslash}p{1.4cm}>{\centering\arraybackslash}p{1.4cm}>{\centering\arraybackslash}p{1.4cm}>{\centering\arraybackslash}p{1.4cm}>{\centering\arraybackslash}p{1.4cm}}
\toprule
\multirow{3.5}{*}{Method}& \multicolumn{5}{c}{\raisebox{-1.5ex}{Scan2Cap (val)}} & \multicolumn{4}{c}{\raisebox{-1.5ex}{ScanQA (val)}} & SQA3D (test) \\
\cmidrule(lr){2-6} \cmidrule(lr){7-10} \cmidrule(lr){11-11}
& CIDEr & BLEU-4 & METEOR & ROUGE & Sim & CIDEr & BLEU-4 & METEOR & ROUGE & EM@1 \\
\midrule
Baseline & 72.4 & 38.2 & 27.9 & 58.1 & 55.3 & 101.4 & 13.2 & \textbf{20.0} & \textbf{49.2} & 50.0 \\
Baseline +10\% RandFilt &  62.7&  40.2&  28.7&  63.2&  57.5&  98.2&  \textbf{15.4}&  19.2&  46.3&  62.5\\
Baseline +20\% RandFilt &  55.4&  38.6&  28.1&  62.8&  56.1&  93.2&  14.3&  18.3&  44.8&  68.4\\
Baseline +30\% RandFilt &  53.7&  38.7&  27.8&  62.7&  54.3&  89.4&  13.7&  17.5&  43.4&  66.6\\
Baseline +40\% RandFilt &  52.3&  38.0&  27.8& 62.3& 55.2& 90.1& 13.4& 18.1& 43.7& 68.9\\
Baseline +50\% RandFilt &  48.0&  32.6&  25.5&  59.7&  47.2&  85.5&  11.9&  15.4&  37.9&  67.4\\
Baseline + QGAP  &  \textbf{78.4}& \textbf{44.1}& \textbf{30.4}& \textbf{65.6}& \textbf{64.2}& \textbf{101.6}& 14.1& 19.9& 48.7& \textbf{70.1}\\
\bottomrule
\end{tabular}
}
\end{table*}

\renewcommand{\arraystretch}{1}
\begin{table*}[htpb!]
\vspace{-1mm}
\centering
{\setlength{\tabcolsep}{6pt}
\caption{
Performance of 3D-SceneQ under simulated camera calibration noise. 
Mild noise perturbs focal length and SE(3) parameters with Gaussian noise sampled from $\mathcal{N}(0, (0.05x)^2)$, and is applied with 10\% probability. 
Severe noise increases the perturbation scale to $\mathcal{N}(0, (0.10x)^2)$ with 15\% probability, where $x$ denotes the original parameter value.
}
\label{tab:mofe}
\resizebox{1\linewidth}{!}{
\begin{tabular}{l>{\centering\arraybackslash}p{1.4cm}>{\centering\arraybackslash}p{1.4cm}>{\centering\arraybackslash}p{1.4cm}>{\centering\arraybackslash}p{1.4cm}>{\centering\arraybackslash}p{1.4cm}>{\centering\arraybackslash}p{1.4cm}>{\centering\arraybackslash}p{1.4cm}>{\centering\arraybackslash}p{1.4cm}>{\centering\arraybackslash}p{1.4cm}>{\centering\arraybackslash}p{1.4cm}}
\toprule
\multirow{3.5}{*}{Method}& \multicolumn{5}{c}{\raisebox{-1.5ex}{Scan2Cap (val)}} & \multicolumn{4}{c}{\raisebox{-1.5ex}{ScanQA (val)}} & SQA3D (test) \\
\cmidrule(lr){2-6} \cmidrule(lr){7-10} \cmidrule(lr){11-11}
& CIDEr & BLEU-4 & METEOR & ROUGE & Sim & CIDEr & BLEU-4 & METEOR & ROUGE & EM@1 \\
\midrule
Baseline & 72.4 & 38.2 & 27.9 & 58.1 & 55.3 & 101.4 & 13.2 & 20.0 & \textbf{49.2} & 50.0 \\
Mild Noise & \textbf{78.0}& 43.4& 30.4& \textbf{65.4}& 63.9& 100.4& 13.8& 19.5& 47.6& \textbf{68.7}\\
Severe Noise & 77.4& 43.6& 30.5& 64.2& 64.1& 98.2& 14.2& 18.9& 46.2& 68.1\\
Baseline + MOFE  & 77.7 & \textbf{44.4} & \textbf{30.5} & 65.1 & \textbf{64.1} & \textbf{102.0} & \textbf{14.6} & \textbf{20.0} & 48.8 & 66.7 \\
\bottomrule
\end{tabular}
}}
\end{table*}

\renewcommand{\arraystretch}{1.1}
\begin{table*}[htpb!]
\centering
{\setlength{\tabcolsep}{6pt}
\caption{Performance of 3D-SceneQ under different training setups. 
\texttt{w/o Align} trains the model without the alignment stage; 
\texttt{ScanNet-only} and \texttt{3RScan-only} are variants trained exclusively on scenes from their respective datasets.
Exact-Match (EM) accuracy is reported for question-answering tasks, while METEOR is used for all other tasks. 
\underline{Underlined values} denote zero-shot results on unseen scenes.}
\label{tab:ablations_data}
\begin{tabular}{l>{\centering\arraybackslash}p{1.5cm}>{\centering\arraybackslash}p{1.5cm}>{\centering\arraybackslash}p{1.5cm}>{\centering\arraybackslash}p{1.5cm}>{\centering\arraybackslash}p{1.5cm}>{\centering\arraybackslash}p{1.5cm}}
\toprule
\multirow{2.5}{*}{Method}& \multicolumn{3}{c}{\texttt{ScanNet}} & \multicolumn{3}{c}{\texttt{3RScan}} \\
\cmidrule(lr){2-4} \cmidrule(lr){5-7}
& Scan2Cap & ScanQA & SQA3D  & 3RQA & 3RDialog & 3RPlan  \\
\midrule
\texttt{w/o Align} &  28.3&  20.4&  \textbf{68.9}&  46.4&  29.8& 21.3 \\
\texttt{ScanNet-only} &  30.4&  22.2&  50.1&  \underline{37.3}&  \underline{6.3}&   \underline{1.9}\\
\texttt{3RScan-only} &  \underline{15.8}&  \underline{14.0}&  \underline{25.9}&  \textbf{50.2}&  29.9&   21.4\\
3D-SceneQ &  \textbf{30.5}&  \textbf{24.3}&  67.0&  48.8&  \textbf{29.9}&  \textbf{22.7}\\
\bottomrule
\end{tabular}
}
\vspace{-3mm}
\end{table*}

\subsubsection{Impact of Filtering Strategies and Ratios in QGAP} To evaluate the effectiveness of QGAP's learnable latent query mechanism, we conduct controlled experiments replacing the adaptive pruning module with two fixed-ratio strategies: LowSimFilt, which discards tokens with the lowest BERT-based cosine similarity to the query, and RandFilt, which randomly removes tokens. We vary the pruning ratio from 10\% to 50\% in 10\% increments and report results in Table~\ref{tab:filt1} and Table~\ref{tab:filt2}.
As shown in both tables, all fixed-ratio pruning strategies result in consistent and monotonic degradation in performance as the pruning ratio increases. This suggests that static heuristics are inherently brittle, as they lack the adaptivity needed to tailor pruning decisions to different query demands: over-pruning risks eliminating crucial context, while under-pruning retains distracting or irrelevant tokens. 
In contrast, QGAP adaptively predicts a query-specific pruning ratio via a learnable latent query mechanism. This allows the model to dynamically retain just the right amount of visual context for each question, leading to superior and stable performance across all benchmarks. These results empirically validate the conceptual motivation behind QGAP: different questions require vastly different levels of contextual detail, general scene descriptions need broader visual coverage, while targeted questions require concise yet relevant context, thereby reducing the influence of redundant information. The learnable latent query enables the model to align token retention with the intent and scope of the query, thereby achieving better performance and robustness than any fixed-rule filtering strategies.

\subsubsection{Robustness of MOFE to Sensor Calibration and Registration Errors}
MOFE relies on accurate 2D-to-3D feature projection, which depends on intrinsic and extrinsic camera parameters. In real-world scenarios, small calibration or registration errors are inevitable. To assess robustness, we simulate two levels of parameter noise during training: 
Mild noise adds Gaussian perturbations (10\% probability) to focal length and SE(3) parameters, sampled from $\mathcal{N}(0, (0.05x)^2)$, where $x$ is the original parameter value. Severe noise increases the probability to 15\% and the noise scale to $\mathcal{N}(0, (0.10x)^2)$.
As shown in Table~\ref{tab:mofe}, model performance remains strong under both conditions. For example, Scan2Cap CIDEr scores reach 78.0 (mild) and 77.4 (severe), close to the clean setup (77.7). On SQA3D, EM@1 slightly improves (68.7 and 68.1 vs. 66.7). This robustness is attributed to the design of MOFE. Instead of relying on pixel-level precision, MOFE aggregates 2D features across full object masks, making it less sensitive to small spatial misalignments. Minor projection errors caused by pose noise typically affect only a few peripheral pixels, which are averaged out during object-level pooling. Furthermore, the public datasets used for evaluation, such as ScanNet and 3RScan, offer highly accurate camera poses, with real-world deviations far smaller than our simulated noise. These results demonstrate that 3D-SceneQ is robust to moderate calibration errors, and MOFE remains effective under realistic conditions.

\begin{figure}[t]
\centering
    \includegraphics[width=1\linewidth]
    {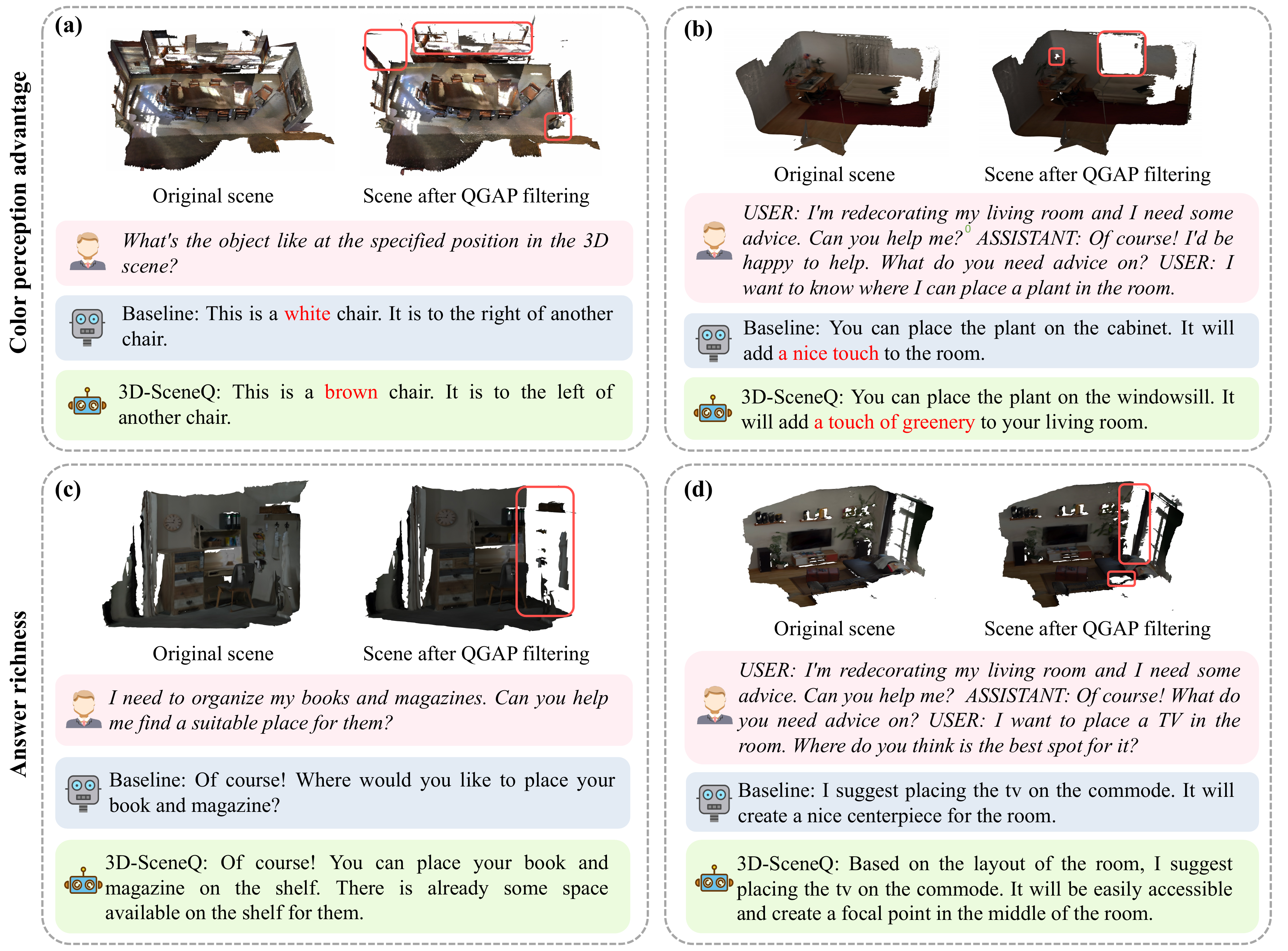} 
    \vspace{-3mm}
    \caption{
   Qualitative comparison between the baseline and our 3D-SceneQ model, with original scenes and their filtered counterparts shown before and after QGAP. (a) and (b) highlight the improved accuracy of 3D-SceneQ's color perception, while (c) and (d) showcase its richer, more detailed answers. Red boxes indicate regions where objects have been pruned by QGAP, distinguishing them from the original scene.
    }
    \label{fig:4}
    \vspace{-3mm}
\end{figure}

\subsubsection{Influence of Training Setups} We analyze the impact of different training setups on 3D-SceneQ performance in Table~\ref{tab:ablations_data}, including variants trained without alignment (\texttt{w/o Align}) and models trained exclusively on ScanNet (\texttt{ScanNet-only}) or 3RScan (\texttt{3RScan-only}). The \texttt{ScanNet-only} variant slightly underperforms compared to the full 3D-SceneQ model even on ScanNet tasks and notably struggles with cross-domain tasks (3RQA, 3RDialog, 3RPlan). Similarly, \texttt{3RScan-only} achieves strong performance on 3RScan tasks but shows limited zero-shot generalization capability on ScanNet tasks. These results underscore the benefits of joint training, which offers broader coverage across diverse scenes and tasks and thereby leads to superior overall performance and enhanced cross-domain generalization in 3D-SceneQ.

\subsection{Visual Analysis}
In Fig.~\ref{fig:4}, we evaluate the performance of 3D-SceneQ against the baseline method across two aspects of 3D scene understanding. 
\textbf{First, 3D-SceneQ demonstrates a significant advantage in color perception.} By fusing rich appearance cues extracted from a large-scale BERT model with the extensive spatial context captured by its 3D backbone, our method can discern subtle chromatic variations that remain unrecognized by the baseline. For example, in Fig.~\ref{fig:4}(a), the baseline misclassifies a brown chair as white, whereas 3D-SceneQ correctly identifies its actual color. Similarly, in Fig.~\ref{fig:4}(b), 3D-SceneQ accurately identifies the green color of the plants, highlighting its improved understanding of color information compared to the baseline. 
\textbf{Second, 3D-SceneQ outperforms the baseline in contextual reasoning.} As shown in Fig.\ref{fig:4}(c), 3D-SceneQ suggests putting books and magazines on shelves or on the tabletop, choices that match everyday usage patterns and ergonomic reach. The baseline, by contrast, offers no specific placement at all. This difference is even more pronounced in Fig.\ref{fig:4}(d), where 3D-SceneQ grounds its recommendation in the room's geometry, selecting the commode because that spot is easy to reach and naturally anchors the room's visual focus. Although the baseline suggests the same location, it does so in a formulaic way, without any spatial justification, highlighting its lack of scene-aware reasoning.

\begin{figure}[t]
\centering
    \includegraphics[width=1\linewidth]
    {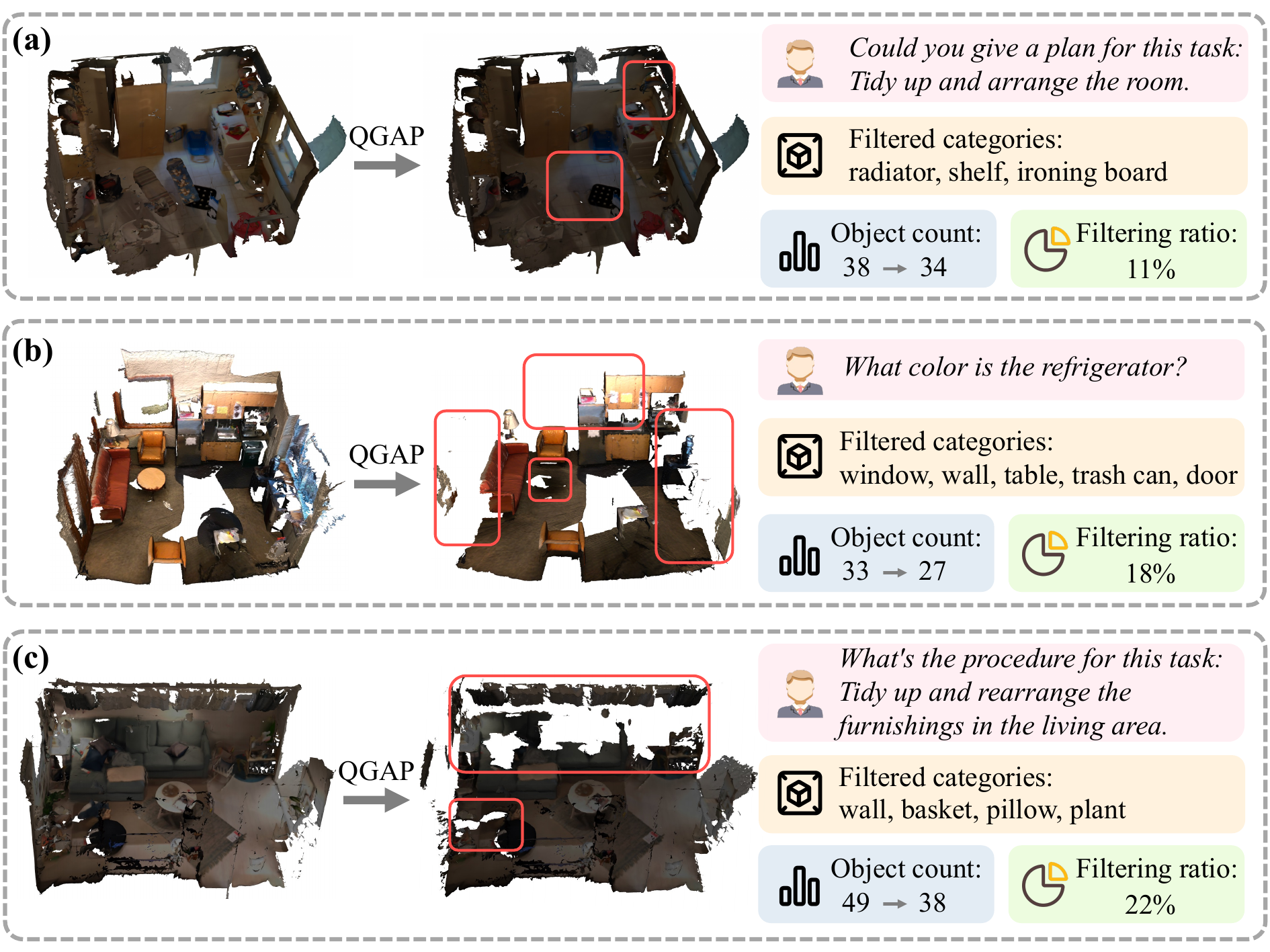} 
    \vspace{-3mm}
    \caption{Adaptive object filtering with Query-Guided Adaptive Pruning (QGAP): (a)-(c) visualize progressively higher pruning ratios, from low to high, revealing how QGAP selectively removes irrelevant objects in response to the user's query. Red boxes indicate regions where objects have been pruned by QGAP, distinguishing them from the original scene.}
    \label{fig:5}
    \vspace{-2mm}
\end{figure}

To evaluate QGAP's scene-pruning strategy, we conduct a series of experiments and visualize three representative scenarios in Fig.~\ref{fig:5}, spanning pruning levels from minimal to aggressive.
In the first example (Fig.\ref{fig:5}(a)), the generic instruction ``Tidy up and arrange the room.'' does not reference any specific object; consequently, QGAP eliminates only 11\% of the object tokens, retaining almost all objects to avoid discarding potentially relevant information.
When the query targets a particular object or category, QGAP applies a more aggressive filter to remove irrelevant elements. For instance, in the second example (Fig.\ref{fig:5}(b)), the query ``refrigerator'' prompts QGAP to drop unrelated items such as windows and walls, resulting in a filtering ratio of 18\%. In the third example (Fig.\ref{fig:5}(c)), the query ``furnishings'' leads to the removal of objects like plants and pillows, raising the ratio to 22\%.
Overall, these results show that QGAP adaptively modulates its pruning strategy to match the semantic intent of each query, enabling context-aware scene refinement that balances preserving crucial content against excising irrelevant tokens.

\begin{figure*}[t]
\centering
    \vspace{-4mm}
    \includegraphics[width=0.98\linewidth]
    {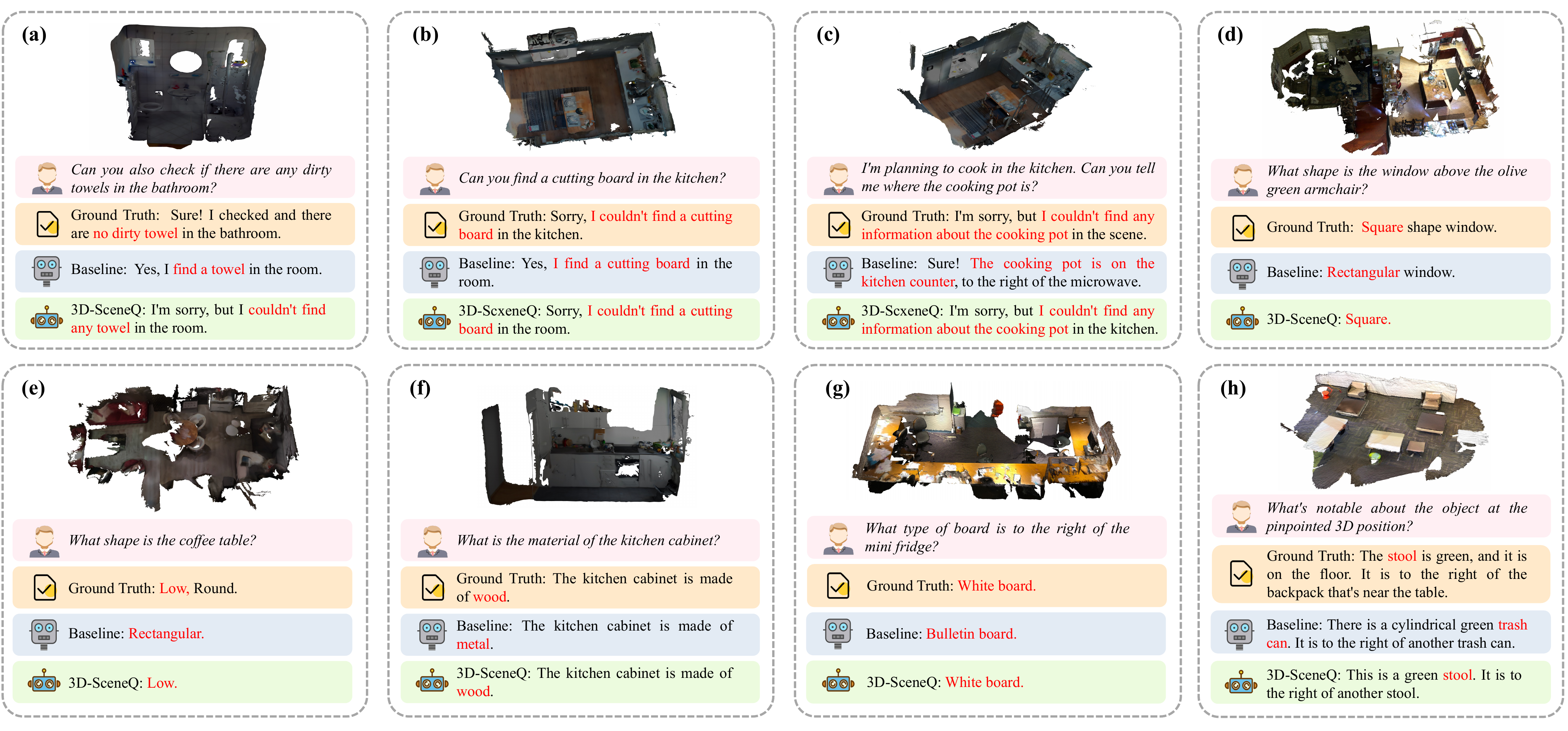} 
    \vspace{-1mm}
    \caption{Qualitative evidence of 3D-SceneQ's hallucination mitigation: (a)--(c) reduce hallucinations of nonexistent or irrelevant objects, (d)--(f) lessen attribute hallucinations, and (g)--(h) correctly distinguish objects with similar visual features. }
    \label{fig:6}
    \vspace{-3mm}
\end{figure*}

\begin{figure*}[h]
\centering
    \includegraphics[width=1\linewidth]
    {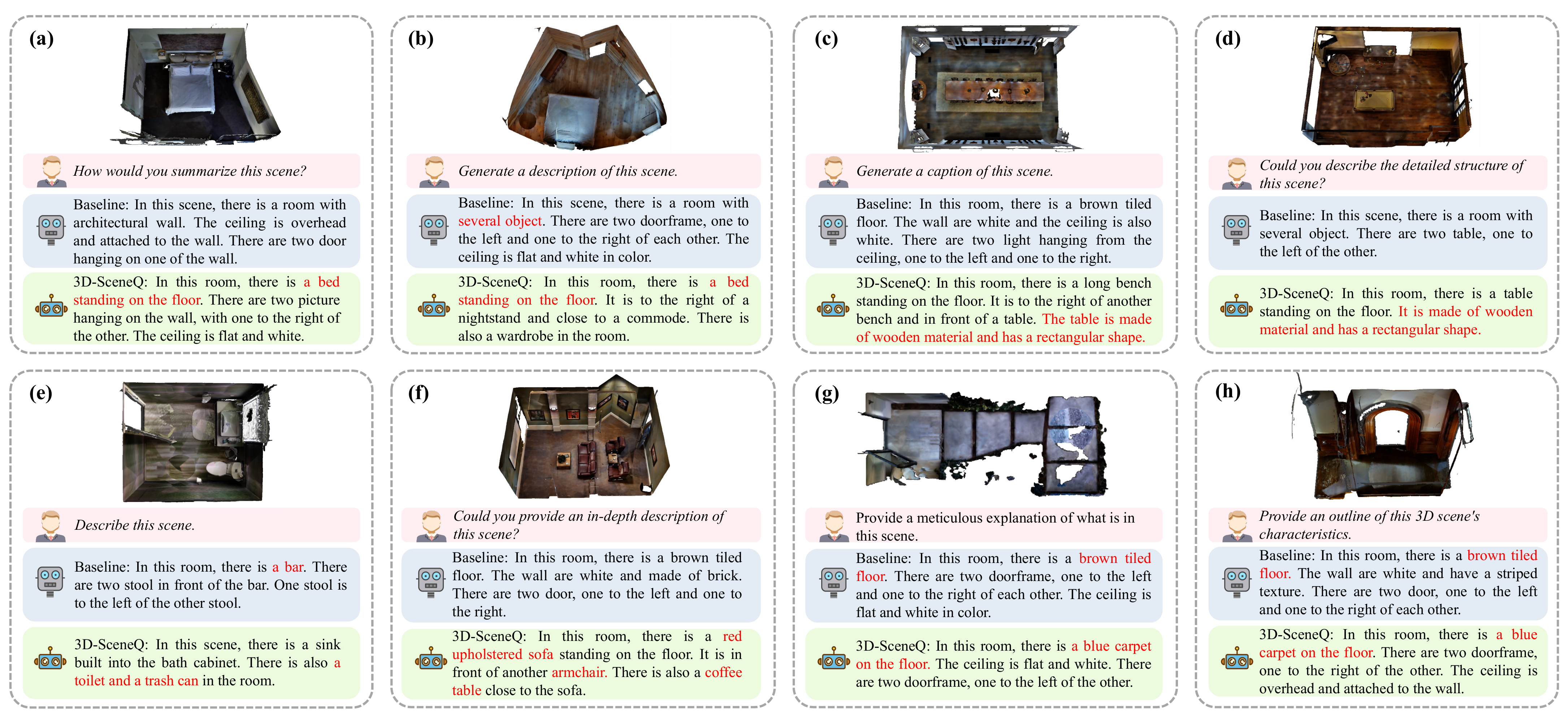}
    \vspace{-5mm}
    \caption{Qualitative comparisons on out-of-distribution scenes from Matterport3D. (a) and (b) demonstrate 3D-SceneQ's capability to identify and describe the most salient objects within the scene. (c) and (d) highlight the model's ability to perceive material and shape attributes. (e) and (f) illustrate the accuracy of object grounding in visually complex environments. (g) and (h) showcase 3D-SceneQ's sensitivity to object color.}
    \label{fig:7}
\end{figure*}

To evaluate 3D-SceneQ's effectiveness in mitigating hallucinations, we carried out a series of visual analyses (Fig.~\ref{fig:5}). The results indicate that our method markedly reduces hallucinations commonly afflicting 3D LLMs, from which we derive three main observations:
\textbf{First, 3D-SceneQ reduces hallucinations of irrelevant or nonexistent objects.}
Unlike the baseline, which invents items absent from the scene, such as a ``dirty tower'' (Fig.~\ref{fig:6}(a)), a ``cutting board'' (Fig.~\ref{fig:6}(b)), or a ``cooking pot'' (Fig.~\ref{fig:6}(c)), 3D-SceneQ always remains faithful to the scene content.  This robustness stems from the collaborative mechanisms of QGAP and MOFE. When the queried object is absent, QGAP adaptively reduces the pruning ratio, as the similarity between all object-category embeddings in the scene and the question embedding is low, thereby preventing forced matches. Meanwhile, MOFE enriches object understanding with multi-view image features, offering stronger semantic cues. Together, they help 3D-SceneQ recognize invalid queries and respond cautiously instead of hallucinating incorrect answers.
\textbf{Second, 3D-SceneQ substantially mitigates perceptual errors in recognizing object attributes.} Hallucinations are not limited to whole objects, they also distort geometry and material. In Fig.~\ref{fig:6}(d)--(f) the Baseline warps a square window into a triangle, morphs a round coffee table into a rectangle, and turns wooden cabinets metallic. In contrast, 3D-SceneQ consistently preserves accurate object properties, maintaining robust attribute recognition.
\textbf{Third, 3D-SceneQ excels at distinguishing objects with similar visual features.} Objects with similar silhouettes or textures often confound vision-language models. The Baseline mislabels a whiteboard as a bulletin board (Fig.~\ref{fig:6}(g)), and a stool as a trash can (Fig.~\ref{fig:6}(h)). 3D-SceneQ consistently delivers reliable responses, demonstrating an exceptional ability to distinguish genuine scene elements from hallucinations. Across a wide spectrum of challenging scenes, 3D-SceneQ consistently delivers precise visual perception and contextual reasoning, establishing a new benchmark for 3D scene understanding through its robust and reliable performance.

To evaluate generalization beyond the training domain, we perform zero-shot experiments on the Matterport3D dataset~\cite{chang2017matterport3d}, a large-scale 3D benchmark containing scenes that 3D-SceneQ has not encountered during training. As it lacks language annotations for captioning or QA, we manually construct a set of prompts to test the model's ability to understand novel scenes. Fig.~\ref{fig:7}(a) and (b) show its ability to identify salient objects, such as correctly detecting the bed at the center, which the baseline misses. In Fig.~\ref{fig:7}(c) and (d), it shows clear awareness of material and shape, capturing fine-grained surface and geometry information. Fig.~\ref{fig:7}(e) and (f) highlight its robustness in grounding objects in cluttered scenes, successfully locating items like a toilet or coffee table. Finally, Fig.~\ref{fig:7}(g) and (h) illustrate accurate color perception, with 3D-SceneQ correctly identifying the blue flooring while the baseline erroneously predicts brown. These results indicate that 3D-SceneQ generalizes well to previously unseen environments, even without fine-tuning, confirming its robustness and broader applicability across domains.

\section{Limitations and Future Work}
\label{sec:limitations}

3D-SceneQ is optimized for indoor scenes and therefore exhibits limited generalization to outdoor settings. Moreover, the absence of explicit temporal modeling reduces robustness in dynamic scenes with moving agents, and its reasoning depth remains constrained without post-training signals. Future work includes adopting more general-purpose encoders and curating balanced indoor--outdoor datasets to narrow the performance gap and strengthen cross-domain generalization. In parallel, the incorporation of post-training signals (e.g., chain-of-thought supervision and reinforcement learning) and the integration of explicit temporal modeling will be explored to enhance reasoning depth and reliability in dynamic scenes.

\section{Conclusion}
\label{sec:conclusion}

In this paper, we present 3D-SceneQ, a novel 3D LLM framework designed to advance scene understanding, reasoning, and planning within complex 3D environments.  To tackle the challenges of redundant object interference and hallucination issues in 3D LLMs, we introduce QGAP, an adaptive pruning mechanism that selectively eliminates irrelevant object tokens. Additionally, to overcome the semantic limitations inherent in point cloud data, we develop MOFE, a multi-modal feature integration module to overcome the semantic limitations of point cloud data by integrating 2D features to enhance the model's comprehension of 3D scenes. Extensive experimental evaluations on benchmark tasks, including 3D captioning, question answering, and task planning, consistently demonstrate the superior performance of 3D-SceneQ compared to SOTA approaches. These results highlight the framework's promising generalization capabilities and enhanced reasoning abilities in challenging 3D environments. Overall, 3D-SceneQ represents a significant advancement in various 3D understanding tasks, offering a strong foundation for future research and applications in intelligent interaction with complex 3D environments.

\bibliographystyle{IEEEtran}  
\bibliography{refs}

\vfill

\end{document}